\documentclass[sigconf]{acmart}
 
\setcopyright{none}
\renewcommand\footnotetextcopyrightpermission[1]{}

\usepackage{graphicx}
\usepackage[font=small,labelfont=bf]{caption}

\usepackage{enumitem}

\usepackage{makecell}
\usepackage{multirow}
\usepackage{url}
\usepackage{xspace}
\usepackage{makecell}

\newcommand{\dll}{DysLexLens\xspace}

\begin{document}

\title{DysLexLens: A Low-Resource LLM Framework for Analysing Dyslexic Learners’ Insights from Online Forums}

\author{Dana Rezazadegan}
\authornote{Corresponding author.}

\author{Atie Kia, Phongpadid Nandavong, Dominique Carlon, Jeremy Nguyen, Abhik Banerjee, James Marshall, Anthony McCosker, Yong-Bin Kang}

\affiliation{%
  \institution{Swinburne University of Technology}
  \city{Melbourne, VIC}
  \country{Australia}
}

\email{{drezazadegan, akia, bnandavong, dcarlon, jdnguyen, abanerjee, jgmarshall, amccosker, ykang}@swin.edu.au}

\newcommand{\yb}[1]{\textcolor{red}{\small\textbf{[YB]} #1 $\triangleleft$}}
\newcommand{\dana}[1]{\textcolor{blue}{\small\textbf{[DR]} #1 $\triangleleft$}}

\begin{abstract}
  Dyslexic learners increasingly use artificial intelligence (AI) tools to support reading, writing, organisation, and study-related tasks. However, their lived experiences with these tools remain largely underexamined. This paper proposes DysLexLens, a low-resource LLM framework, designed to analyse dyslexic learners’ experience with AI through online forum discussions. DysLexLens is designed as an end-to-end, evidence-traceable architecture which transforms noisy social media posts into a dictionary-driven corpora, provides knowledge-graph (KG)-based question reasoning, generates verifiable query responses, and enables response evaluation through quantitative and human-grounded assessment. DysLexLens has four key features. First, it employs a dictionary-driven filtering method to construct a more focused Reddit corpus on dyslexia and AI, filtering out noisy and weakly related posts to improve the relevance of data collected from low-resource forum contexts. Second, it integrates LLM-assisted semantic analysis with KG-based query reasoning to uncover meaningful patterns. Third, it has quantitative evaluation metrics (RAGAS and Query Robustness) to measure LLM-generated response performance. Fourth, it provides structured qualitative validation guidelines for assessing response quality, with a specific focus on hallucination and evidence alignment. We demonstrate the effectiveness of DysLexLens using dyslexia-related Reddit forum data and 30 questions. The results show its potential generalisability to other low-resource forum data contexts. \dll, sample data, questions and evaluation results are available at Github\footnote{https://github.com/SIRI-HAC-Program/DysLexLens} to support reproducibility.
    
\end{abstract}


\maketitle

\section{Introduction}

Dyslexia is widely recognised as a neurobiological learning difficulty that affects accurate and fluent word reading, spelling, and decoding, despite adequate intelligence, motivation, and educational opportunity~\cite{lyon2003definition}. Its underlying causes are multifactorial, encompassing neurobiological, psychological, and environmental influences~\cite{catts2024revisiting}. Recent prevalence estimates suggest that dyslexia affects 5\%-17\% of the population due to reading difficulty, the discrepancy between expected and observed reading performance, and some literacy-related measures used in assessment~\cite{wagner2020prevalence}.
Learners with dyslexia often face important educational challenges, such as difficulties in reading fluency, written communication, comprehension, organisation, and classroom participation. These challenges create a strong need to better understand the forms of support, coping strategies, and practical solutions to improve their learning outcomes. 

The rapid rise of AI tools has introduced new possibilities for personalised and accessible learning support for dyslexic learners.
To help dyslexic learners, existing works have primarily focused on two directions. The first is technology-centred, including reviews of assistive technologies for dyslexic learners~\cite{smith2020assistive,lerga2021review}. While valuable, this direction primarily maps available tools and functions, offering limited insight into how dyslexic learners themselves describe their needs, frustrations, and lived experiences in naturalistic settings. The second direction is learner-centred, examining lived experience and online discourse. For example, Reddit-based thematic analysis has shown that online communities can reveal meaningful self-expressed perspectives from people with dyslexia~\cite{pub:65001}. However, this direction focuses more on identity construction and social meaning, rather than AI perceptions and learning-related use cases. Thus, prior research explains either the supply side of dyslexia support technologies or selected aspects of dyslexic experience, but provides limited learner-centred evidence on how AI and related tools are perceived, adopted, and evaluated in practice.

To address this gap, this paper proposes, \dll, an evidence-traceable framework for analysing Reddit forum data of dyslexic learners and related stakeholders. \dll helps to investigates how AI tools are discussed in naturalistic online communities, with a focus on learning-related practices, perceived benefits and limitations, enabling conditions, educational support contexts, and temporal shifts in discourse. Specifically, in the paper, we will demonstrate how \dll can addresses five research questions: \textbf{RQ1:} What learning-related use cases for AI tools are described by dyslexic learners in online discussions? \textbf{RQ2:} What benefits and failure modes are reported when AI tools are used for different learning tasks? \textbf{RQ3:} Under what conditions are AI tools perceived as helpful, supportive, or inclusive for dyslexic learners? \textbf{RQ4:} How are AI-based supports discussed in relation to broader educational challenges, including institutional accommodations and personal coping strategies? \textbf{RQ5:} How have discussions of AI tools among dyslexic learners changed over time in terms of prevalence, tone, and perceived role in learning support?

This paper makes the following contributions:
\begin{itemize}
  \item We propose \dll, designed to be a generalisable, evidence-traceable framework for analysing low-resource forum data, where relevant discussions are sparse, noisy, and difficult to identify at scale.

  \item We demonstrate the effectiveness of \dll in the domain of dyslexia and AI, using Reddit discussions to examine how dyslexic learners and related stakeholders describe AI-related learning use cases, benefits, limitations, and support needs.

  \item We present an LLM-based analysis pipeline of \dll that links generated responses to supporting corpus evidence through KG-based reasoning.

  \item We provide a hybrid evaluation approach combining quantitative metrics and human-grounded validation guidelines to assess response quality, hallucination and evidence alignment.
\end{itemize}

\section{Related Work} \label{sec:related_work}
This section positions the study at the intersection of dyslexia support, AI-enabled learning technologies, and LLM-assisted analysis of online forum data. We first review research on dyslexic learners and existing support approaches, before discussing the use of LLMs for analysing user-generated forum data.

\textbf{Research on Dyslexic Learners:}
A large body of research has examined dyslexic learners,  extending beyond decoding and spelling difficulties \cite{snowling_defining_2020} to include comprehension \cite{georgiou_meta-analytic_2022}, written expression \cite{graham_children_2021}, learner participation \cite{nevill_social_2023}, learner confidence \cite{hamiltonclark_dyslexia_2024}, and wider classroom experience \cite{ross_im_2021}.  
Existing work on support for dyslexic learners can be grouped into three broad strands: non-technological pedagogical interventions, non-AI technological supports, and AI-specific approaches.

The first strand has examined pedagogical interventions, such as phonics-based reading instruction and structured spelling programs. Evidence is not always specific to dyslexia and is sometimes drawn from studies of broader categories, such as reading disabilities. A meta-analysis of randomised controlled trials found that phonics was the only reviewed approach that produced statistically significant effects on the reading and spelling ability of children and adolescents with reading disabilities \cite{galuschka_effectiveness_2014}. A more recent review also found empirical support for orthographic and morphological instruction \cite{galuschka_effectiveness_2020}. However, support for dyslexic learners is not only a matter of literacy outcomes. Pedagogical choices can also shape whether learners feel understood, whether they feel singled out as different, and the extent to which they have agency over how their dyslexia is supported \cite{ross_im_2021}.

The second strand has focused on non-AI technological supports, including assistive tools for reading and writing. Supportive software can improve   spelling ability, but it is generally viewed as an aid rather than a substitute for direct instruction \cite{galuschka_effectiveness_2020}. A five-year follow-up study of nine dyslexic students found that continued use of tools such as audiobooks, text-to-speech, and speech-to-text depended onschool-level context, students' emotional responses in the classroom, and whether students developed meaningful strategies for using the tools \cite{almgren2024dyslexic}. These findings indicate that technology availability alone is insufficient. Rather, tools may be difficult to learn, unsuitable for some students' needs, or hard to integrate into existing learning practices~\cite{hamiltonclark_dyslexia_2024}.

The third strand has explored AI-specific interventions. A review work identified four prominent uses of AI for dyslexic learners: early detection and diagnosis, personalised learning, speech and language processing, and neuroimaging \cite{yap_artificial_2025}. Another review work of AI-based interventions for learners with learning disabilities, in which dyslexia was the most frequently studied condition, found positive outcomes but also noted risks of bias in the existing evidence base\cite{paglialunga_effectiveness_2025}. 

These strands show that existing research has largely examined either support interventions, available technologies, or the measured effects of AI-based systems. Less is known about how dyslexic learners themselves discuss AI tools in naturalistic settings, including what they use them for, where they experience benefits or failures, and how they perceive such tools. Our study addresses this gap by analysing dyslexia- and AI-related Reddit forum data as a low-resource, user-generated data setting.

\textbf{LLM-Assisted Online Forum Data Analysis:}
The growing use of LLMs for analysing text data has increased interest in applying them to online forum and social media data. Existing studies have used LLMs for market sentiment analysis \cite{dengllmfinancial}, generating labels for supervised learning models \cite{dengLLMSentiment}, and analysing public sentiment during social movements \cite{sidhartaBERT}. A prominent application area is health-focused online discourse, including Reddit-based analyis of lupus pain narratives \cite{walkerlupus}, dietary and weight-loss discussions \cite{kaloudis2025ai}, and eating disorder discourse \cite{chopra2024deciphering}. 
Mental health communities have also been a major focus of LLM-assisted forum analysis.   Existing work has used LLMs to identify latent linguistic dimensions associated with suicidality in Reddit forums and to analyse linguistic characteristics across social networking posts related to mental health~\cite{bauer2024using,kim2023understanding}. Related research also examines online discussions of LLMs themselves, such as ChatGPT, as emerging mental health support tools~\cite{jungChatGPTmental,luo2025shaping}.

While these studies demonstrate the value of LLMs for analysing online discussions, most focus on broad sentiment, thematic patterns, or mental health signals. Less attention has been given to low-resource forum settings where relevant posts are sparse, noisy, and difficult to identify, and where analytical outputs need to be linked back to supporting evidence. 

\section{DysLexLens Framework} \label{sec:method}

\begin{figure*}[!t]
\centering
\includegraphics[width=6.9in, height=2.7in]{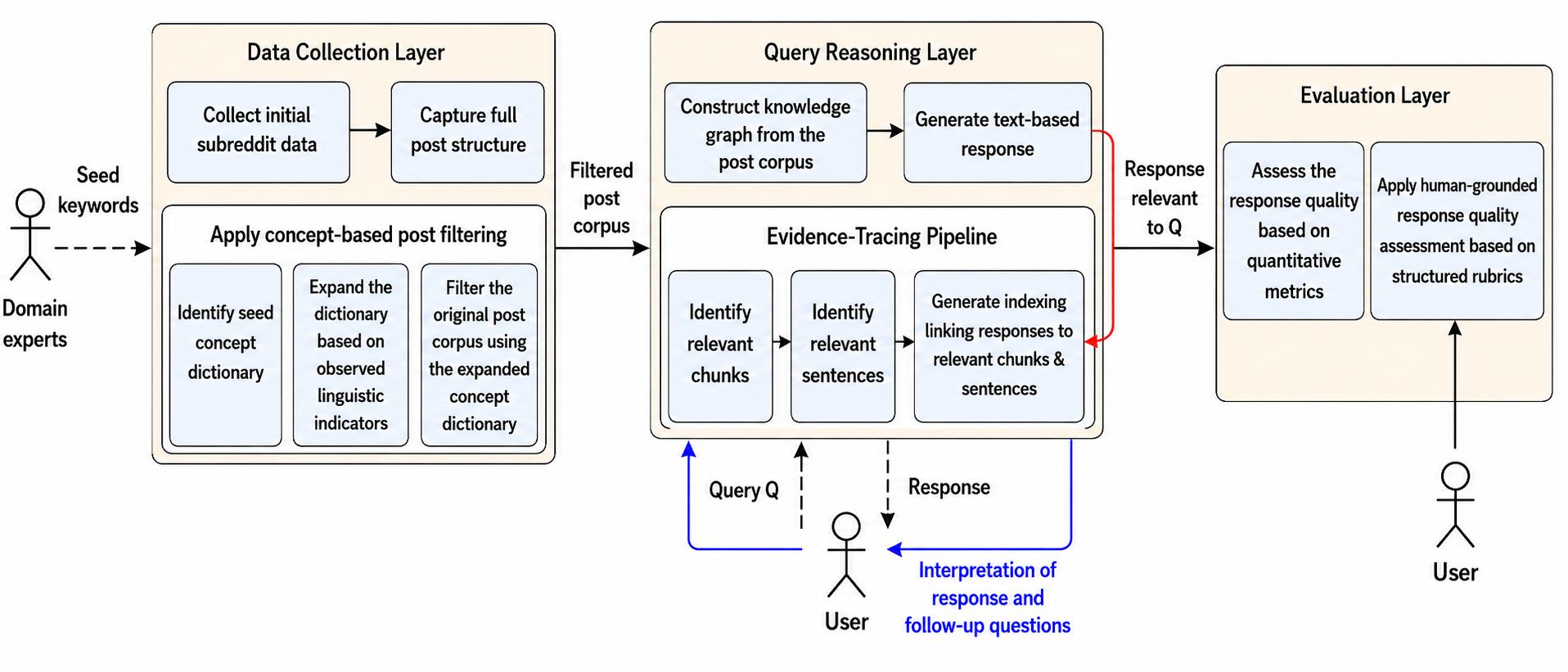}
\vspace{-10pt}
\caption{\textcolor{black}{The overview of \dll framework}}
\label{fig: DysLexLens}
\end{figure*}

This section presents \dll, an evidence-traceable framework for analysing low-resource forum data. \dll is designed for settings where relevant discussions are sparse, noisy, and distributed, making it difficult to construct a focused corpus for systematic analysis. Rather than treating forum data as a directly usable dataset, \dll supports the workflow from targeted data collection to query-based reasoning and response evaluation. In this paper, we apply \dll in the domain of dyslexia and AI by analysing Reddit discussions in which dyslexic learners and related stakeholders describe learning experiences, support needs, and perceptions of AI-supported tools. As seen in Fig.~\ref{fig: DysLexLens}, \dll consists of three layers: data collection, query reasoning, and evaluation.

\subsection{Data Collection Layer}

The goal of this layer is to construct a focused corpus from noisy, low-resource forum data. In the dyslexia and AI case study, relevant discussions are not concentrated in a single subreddit or labelled dataset; instead, they are scattered across communities related to dyslexia, neurodiversity, education, accessibility, assistive technology, and AI tools. To address this challenge, \dll adopts a three-step process: collecting candidate forum data from a broad set of relevant communities, preserving post-comment discussion structure while removing low-information records, and applying a concept dictionary-based filtering method to identify posts most closely aligned with the research objectives.

\textbf{Collect Initial Subreddit Data:} Reddit is selected as the data source because it contains publicly accessible discussions in which individuals with dyslexia, parents, educators, and other stakeholders share experiences, challenges, coping strategies, and views on assistive technologies and AI-supported learning. To identify candidate communities, we first identify seed keywords related to dyslexia, neurodiversity, learning support, assistive technology, and AI from literature studies. 
Using the Arctic Shift API, posts and discussion threads are collected from 50 screened subreddits\footnote{These 50 subreddits are listed in our GitHub repository (see footnote 1).}, including communities related to dyslexia and learning differences, broader neurodiversity, education, accessibility, assistive technologies, AI tools, and general user experience discussions. This broad initial collection is intentional. In low-resource forum settings, narrowly querying only a small number of domain-specific communities risks missing relevant discussions that occur in adjacent communities.

\textbf{Capture Full Post Structure and Remove Noise Posts:} For each subreddit, non-stickied posts are retrieved to capture ordinary user-generated discussions rather than moderator announcements or pinned reference material. For each post, recursive comment extraction is then performed to capture the full discussion structure, including top-level comments and nested replies. Then, the extracted posts and comments are merged into a single datafarme, with subreddit labels derived from the source filenames. The resulting initial corpus contains 23,480 posts and comments, comprising 1,663,250 words across 45 subreddit communities.  suitable for analysing dyslexia-related experiences, support needs,  perceptions of AI and assistive technologies. To reduce low-information noise, posts with fewer than three words are excluded.

\textbf{Concept Dictionary-based Post Filtering:} 
The initial Reddit corpus contains many posts that are only weakly related to the study focus. This is a common challenge in low-resource forum data, where relevant discussions are sparse, noisy, and distributed across multiple communities. To construct a more focused corpus for downstream analysis, \dll applies a concept dictionary-based filtering method.
The dictionary is developed using a construct-driven procedure, aligned with established practices in computational text analysis~\cite{kuckartz2019qualitative,bolden2000bridging}. We first define five core concepts aligned with the research objectives: \textit{dyslexia}, \textit{AI}, \textit{technology}, \textit{learning support}, and \textit{perception}. For each concept, we identify observable linguistic indicators from prior literature, domain expertise, and common expressions used in online forum discussions. For example, the \textit{learning support} concept includes terms related to speech to text, spell checkers, predictive text, reading aids, note-taking, and similar supports, while the AI concept includes terms related to AI, large language models, machine learning, ChatGPT, Claude, Gemini, and related technologies.  

The seed dictionary is then expanded with related terms and close variants. The filtering process first applies exact keyword matching across the full post text. It then uses similarity scoring to identify semantically close variants that may not appear in the original dictionary. For example, \textit{dictation}, \textit{dictating}, \textit{diction}, \textit{voice typing} and \textit{speech-to-text} are added to the \textit{learning support} dictionary as terms related to \textit{speech to text}. Candidate variants are exported for manual review, and validated terms are added back into the dictionary for subsequent filtering and analysis. 

This process reduces the candidate corpus to a focused dataset of 319 posts from 27 subreddits. The resulting dataset is not intended to represent all Reddit discussions about dyslexia. Rather, it provides a targeted, research question-aligned (i.e., \textbf{RQ1}-\textbf{RQ5}) corpus for analysing how dyslexia and AI are discussed in low-resource Reddit forum data.

\subsection{Query Reasoning Layer}

This layer enables analysis of the filtered forum corpus. Its goal is to support analysis by retrieving relevant evidence, reasoning over semantic relations, generating responses to user-given queries (or questions), and linking generated claims back to the original source text. As a prerequisite to query execution, \dll first constructs a knowledge backbone (i.e. KG) from the final filtered corpus produced in the previous layer. This KG is built once and used across all user interactions, rather than being dynamically reconstructed for each query. This layer consists of three components: KG-based retrieval, evidence-grounded response generation, and user-guided follow-up analysis.

\textbf{Prerequisite Knowledge Graph Construction:}
Before query reasoning is performed, \dll constructs a reusable KG from the final filtered corpus. We use the Property Graph Index in LlamaIndex to build the graph representation. The corpus is first divided into text chunks, and an LLM extracts semantic triples from each chunk in the form of subject--predicate--object relations. Subjects and objects are represented as entity nodes, while predicates are encoded as labelled edges. The original text chunks are retained as source nodes so that extracted relations can be traced back to their supporting textual evidence. 
To support query-time retrieval, vector embeddings are generated for indexed chunks and nodes using `{text-embedding-3-small}'. Thus, the KG provides the relational structure, while the embeddings support semantic matching between user queries and relevant corpus evidence.

\textbf{KG-based Retrieval: }
Given a user query $q$, \dll's  retrieval process identifies relevant semantic triples, associated source chunks, and supporting sentences from the corpus. The triples provide relational context by showing how key concepts, tools, needs, and experiences are connected, while the retrieved source chunks provide textual evidence for answer generation.
The retrieved triples and supporting text are then provided to the LLM to generate a text-based response. The response includes inline references to indexed source chunks, allowing users to inspect the evidence behind the generated answer. In this way, this layer combines semantic interpretation from the KG with source-grounded response generation.

\textbf{Evidence Tracing Pipeline:} \dll applies a three-stage evidence-tracing pipeline after response generation. First, it examines the generated response to identify claims, defined as sentences that include explicit references to relevant source-chunk identifiers. Second, it aggregates the chunk identifiers associated with each identified claim, denoted as \texttt{C*}. Third, it uses these identifiers to retrieve the corresponding original Reddit records, including posts or comments, from each linked source chunk. 

\textbf{User-guided Follow-up Analysis:}
\dll supports multi-turn analysis by allowing users to ask follow-up questions over the same KG context, retrieved source chunks, and relevant subgraphs. This enables users to refine, extend, or challenge an initial response without restarting the retrieval process. Instead, the system preserves the analytical context from the previous query and uses it as the basis for deeper exploration. This human-in-the-loop capability is important for low-resource forum analysis, where relevant evidence may be sparse, fragmented, or open to multiple interpretations. It allows users to examine additional nuances, compare alternative explanations, identify evidence gaps, and surface details that may not be captured in the initial response.

\begin{figure*}[!t]
\centering
\includegraphics[width=6.9in, height=2.9in]{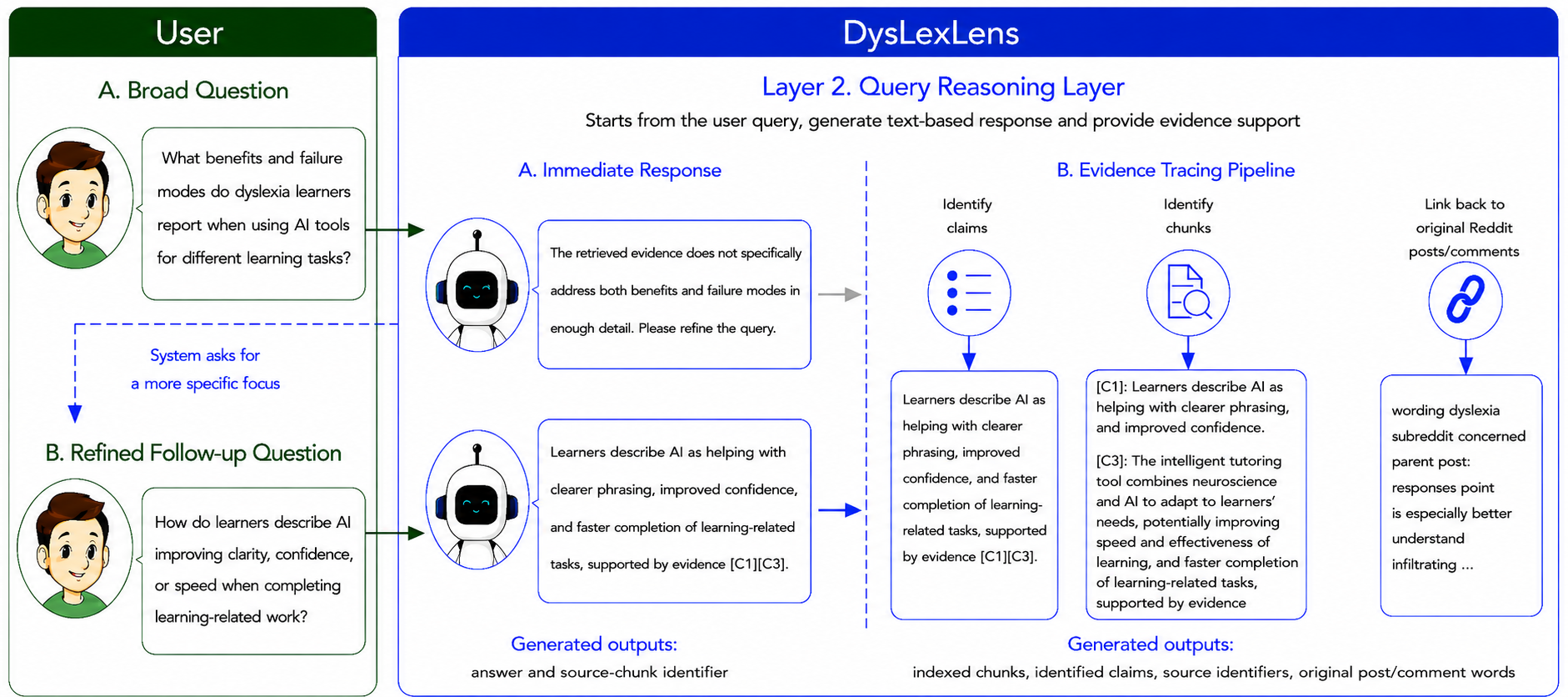}
\vspace{-10pt}
\paragraph{}
\caption{\textcolor{black}{An example of query processing workflow in \dll}}
\label{fig: DysLexLens_Illustrative}
\end{figure*}
\subsection{Evaluation Layer}
This  assesses the quality and evidence alignment of \dll outputs using both quantitative metrics and human-grounded assessment (using structured guidelines). 
The quantitative metrics consist of a RAGAS evaluation metric set and query robustness. The aim is to evaluate not only whether the framework generates relevant responses, but also whether those responses are factually consistent, grounded in retrieved evidence, and interpretable through traceable provenance. 

\textbf{RAGAS Evaluation Metrics:}
Retrieval and generation quality are evaluated using RAGAS evaluation metrics. We construct a test set of 30 queries\footnote{These queries are found at our Github repository.}, consisting of the five research questions (introduced in Section~1 (\textbf{RQ1-5}), each accompanied by five follow-up questions. For each query, \dll generates a response and records the retrieved supporting context, source-chunk index, and source evidence.
Performance is assessed using four metrics: \textit{Answer Relevancy} measuring how well the response addresses the query; \textit{Faithfulness} evaluating factual consistency with the retrieved context; \textit{Context Relevance} assessing the relevance of retrieved text segments; and \textit{Response} \textit{Groundedness} measuring the extent to which generated claims are supported by retrieved evidence. To ensure fair comparison, all final RAGAS results use a fixed retrieval configuration: similarity top-k=3 and a 512-token chunk size. segmentation.

\textbf{Query Robustness Analysis:}
Query robustness analysis evaluates whether \dll produces stable responses when the similar questions are given. This is important because \dll is designed for user-guided analysis rather than fixed-question benchmarking. In low-resource forum data, relevant evidence is often sparse, noisy, and limited. Therefore, small changes in query wording may affect which chunks, subgraphs, or evidence traces are retrieved.
For each research question (\textbf{RQ1-5}), we create three semantically aligned query variants: the original wording, a paraphrased version, and a keyword-based version. All variants are processed using the same experimental configuration. Robustness is assessed by comparing variation in the four RAGAS metrics.
Lower variation indicates stronger robustness, suggesting that \dll can retrieve relevant evidence and generate grounded responses despite differences in query phrasing. Higher variation, particularly in \textit{Context Relevance} or \textit{Response Groundedness}, indicates potential evidence mismatch or instability in response generation.

\textbf{Human Assessment: }

To complement the quantitative metrics, we conduct a human assessment on a purposive sample of generated responses. The sample is selected to include both strong and weak RAGAS score patterns. Humans assess response quality using rubric-based guideline which is developed for this study\footnote{The guidelines are found at our Github repository.}. The rubrics are designed to support consistent assessment of hallucination risk, evidence alignment, and interpretability.

Each sampled response is decomposed into claims, where \texttt{C*} exist. Each claim is then traced to its identified source chunk, and original Reddit posts. The human assesses whether evidence is present, whether the evidence sufficiently supports the claim, and whether the provenance trail is interpretable. Support strength is rated on a 3-scale: \textit{strong, partial, and weak}. This audit helps identify cases where generated responses are well grounded, partially supported, or affected by retrieval mismatch, unsupported inference, or potential hallucination.  

For instance, a response may correctly identify that an AI tool supported writing, while overextending the same evidence to claim improved confidence. The rubric makes this distinction visible through separate assessments of evidence presence, support strength, and interpretative utility. 

\textbf{Illustrative Example}
 
Fig.~\ref{fig: DysLexLens_Illustrative} demonstrates an example of the query-processing workflow in DysLexLens, showing how the framework supports user-driven query refinement, evidence-grounded response generation, and evidence tracing. 

A user may submit any analytical question over the filtered forum corpus, such as \textbf{RQ2}, which asks what benefits and failure modes dyslexic learners report when using AI tools for different learning tasks. If the retrieved evidence does not sufficiently support all aspects of the query, the query reasoning layer returns an immediate response indicating insufficient contextual evidence and enables the user to refine the question.
The user can then give a more specific follow-up question, for example asking how learners describe AI as improving clarity, confidence, or speed when completing learning-related work.  \dll then generates a response with source-chunk identifiers such as \texttt{C1} and \texttt{C3}, which make the evidence trail explicit. Through the evidence-tracing pipeline, these chunck identifiers allow users to identify claims and trace back to the identified source chunks, and the original Reddit records including posts or comments. 

\dll' workflow thus enables users to move from an initial query to more precise, traceable, and interpretable findings while preserving a link between generated responses and the underlying forum evidence.

\section{Evaluation and Results}
\label{sec:evaluation-results}

Our evaluation questions include five research questions and 25 follow-up questions. 
The KG constuction and all responses are generated using gpt-4o-mini.  

\subsection{Evaluation using RAGAS Metrics}
\begin{figure}[!t]
  \begin{center}
  \includegraphics[width=3.2 in, height=3.8in]{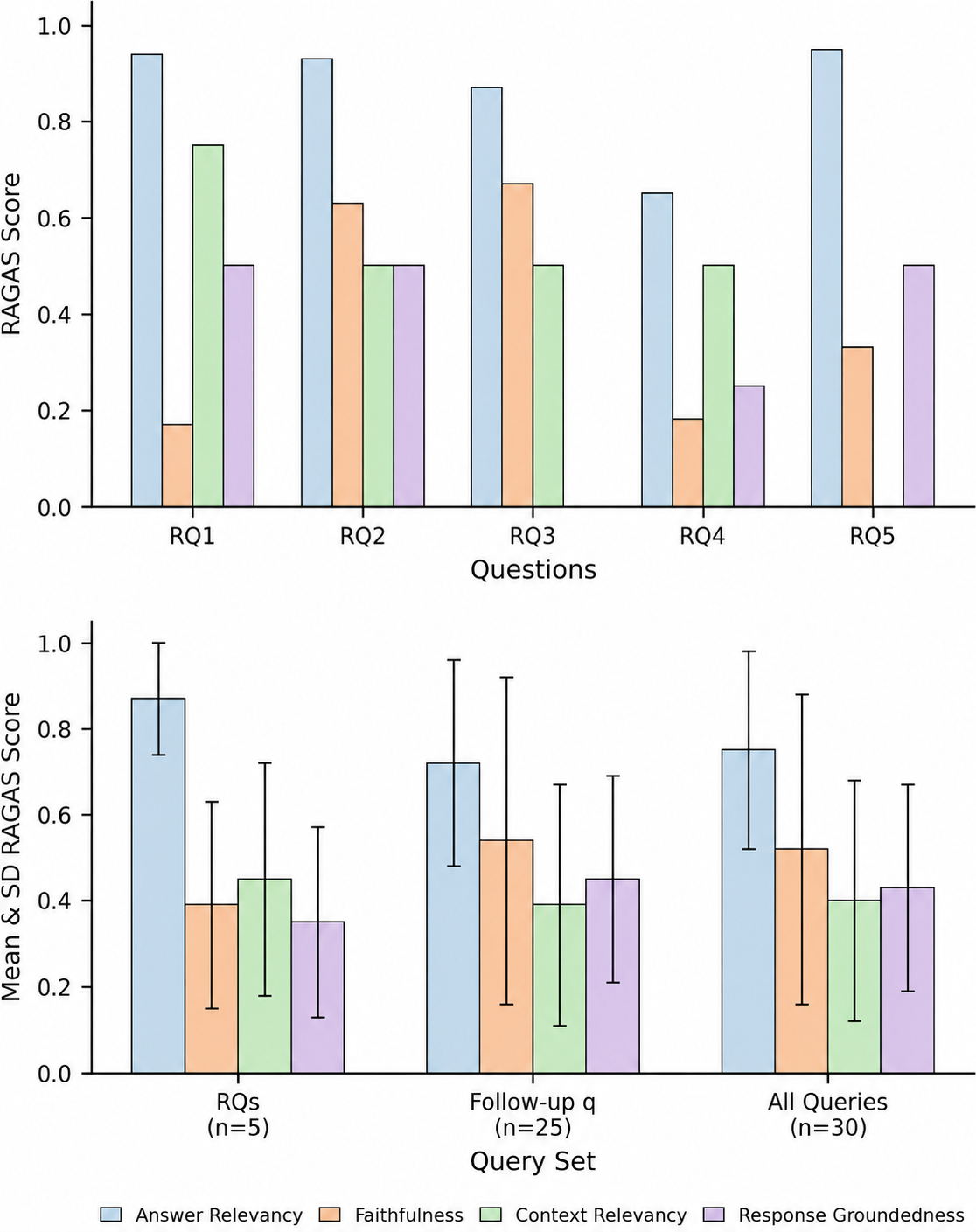}\\
  \caption{Quantitative evaluation results: Single scores for individual \textbf{RQs}; Mean and standard deviation for query set variants.}
  \label{fig: RAGAS}
  \end{center}
\end{figure}

Fig.~\ref{fig: RAGAS} reports the RAGAS results for responses to the five \textbf{RQs}, as well as the mean and standard deviation for the \textbf{RQs}, follow-up questions, and all queries combined. Across all 30 responses, \dll achieves a mean \textit{Answer Relevancy} score of 0.75. In total, 25 of the 30 responses scored at least 0.65 for \textit{Answer Relevancy}. This indicates that most generated responses address the intended meaning of the input questions, although \textit{Answer Relevancy} alone does not show evidential support or factual grounding. 

The research question responses achieve stronger \textit{Answer Relevancy} than follow-up responses, with mean scores of 0.87 and 0.72, respectively. This may suggest that \dll performs better when questions include keywords closely aligned with the concept dictionary. 
However, \textit{Faithfulness}, \textit{Context Relevance}, and \textit{Response Groundedness} for research questions are lower than \textit{Answer Relevancy}.  
Across all 30 queries' responses, the mean \textit{Faithfulness} score is 0.52, mean \textit{Context Relevancy} is 0.40, and mean \textit{Response Groundedness} is 0.43. 

The research question scores further show that \textit{Answer Relevancy} alone does not guarantee evidential support. 
\textbf{RQ1}, \textbf{RQ2}, \textbf{RQ3}, and \textbf{RQ5}  achieve high \textit{Answer Relevancy}, but their evidential support levels differ. For example, \textbf{RQ3} achieves strong \textit{Faithfulness} but a \textit{Response Groundedness} score of zero, suggesting that the response is judged to be factually consistent with the retrieved context, but the generated claims are not explicitly supported by identified evidence. \textbf{RQ5} achieve the highest \textit{Answer Relevancy} score but a \textit{Context Relevancy} score of zero, showing that temporal-change questions are difficult for the retrieval component. 

The evidence-tracing results further provide  an integrated interpretation across both research questions and follow-up queries. Of the 30 generated responses, 29 includes at least one chunk identifier. At the sentence level, the export contained 114 unique claims, of which 96 are linked to at least one source-chunk identifier and 18 have no source-chunk identifier. 

For identified source chunks, the mean retrieval score is 0.86, while identified claims have a mean retrieval score of 0.85, at the response level. However, three identified claims have an average retrieval score of 0.50 or below, showing that retrieval similarity alone is not sufficient for strong claim-level evidential support. 

\subsection{Query Robustness Analysis}
To evaluate query robustness, all 30 evaluation questions are tested using three semantically aligned query formulations: the original question, a paraphrased version, and a keyword-perturbed version. The paraphrased and keyword-perturbed variants are generated using GPT-5.5 Pro. 

The original queries achieve the highest \textit{Answer Relevancy} score (0.75), followed by the paraphrased queries (0.58). This suggests that \dll remains reasonably stable when questions are reworded but their meaning is preserved. However, the keyword-perturbed queries show a clear decrease in \textit{Answer Relevancy} (0.34), indicating that the framework is more sensitive when domain-specific terms are changed.

The highest \textit{Faithfulness} score belongs to the keyword-perturbed query set (0.66), suggesting that these responses are still aligned with the retrieved evidence, despite their lower \textit{Answer relevancy}. This suggests that the generated responses remain relatively consistent with the retrieved evidence, but the retrieved evidence may not match the intended meaning of the original question. In other words, keyword perturbation may lead the system to retrieve narrower or different evidence that can still support the generated response, while reducing alignment with the user’s intended query.

Across all variants, \textit{Context Relevancy} and \textit{Response Groundedness} remain lower than \textit{Answer Relevancy} and \textit{Faithfulness}. The original queries achieve the strongest evidential support, with \textit{Context Relevancy} of 0.40 and \textit{Response Groundedness} of 0.43, while paraphrased and keyword-perturbed queries show lower grounding scores. This indicates that retrieval precision and claim-level evidential support remain the main limitations of the current pipeline.

Hence, \dll is moderately robust to paraphrasing but less stable under keyword perturbation, highlighting the need for improved evidence ranking, and query optimisation guided by core domain-related keywords.

\begin{table}[t]
\centering
\caption{Query robustness across query variants based on mean RAGAS scores. }
\label{tab:query-robustness-mean}
\large
\resizebox{\columnwidth}{!}{%
\begin{tabular}{lcccc}
\toprule
\textbf{Query variant} 
& \textbf{\makecell{Answer\\Relevancy}} 
& \textbf{\makecell{Faithfulness}} 
& \textbf{\makecell{Context\\Relevancy}} 
& \textbf{\makecell{Response\\Groundedness}} \\
\midrule
Original & 0.75 & 0.52 & 0.40 & 0.43 \\
Paraphrased & 0.58 & 0.55 & 0.31 & 0.25 \\
Keyword-perturbed & 0.34 & 0.66 & 0.33 & 0.28 \\
\bottomrule
\end{tabular}%
}
\end{table}

\subsection{Human Assessment}
\label{subsec:provenance-audit-results}

A human assessment is conducted in three dimensions following our structured qualitative assessment guideline to assess the interpretability and verifiability of \dll responses beyond automated RAGAS scores. An evaluation file is automatically generated for this assessment, where multiple rows can appear for a single query response, depending on the number of identified claims, in the response. Each claim in a query response is supported by a \texttt{C*} identifier, which links to the corresponding source chunk. 

Human assessors review 100 claims, including 51 claims with stronger RAGAS metric scores and 49 claims with lower scores, and examine provenance tracing and alignment between generated claims, identified source chunks, and original Reddit records. This assessment is conducted through three audit dimensions: (A) Evidence Verification, (B) Support Strength, and (C) Interpretative Utility. Evidence verification indicates whether the identified chunk appears in the original source. Support strength rates how strongly the identified chunk supports the claim. Interpretative utility rates how useful the identified chunk is for interpretation. Each claim is checked against the exported  \texttt{source\_chunk}, and \texttt{full\_post} fields. 

\begin{figure}[!t]
  \begin{center}
  \includegraphics[width=3.2 in, height=2.04in]{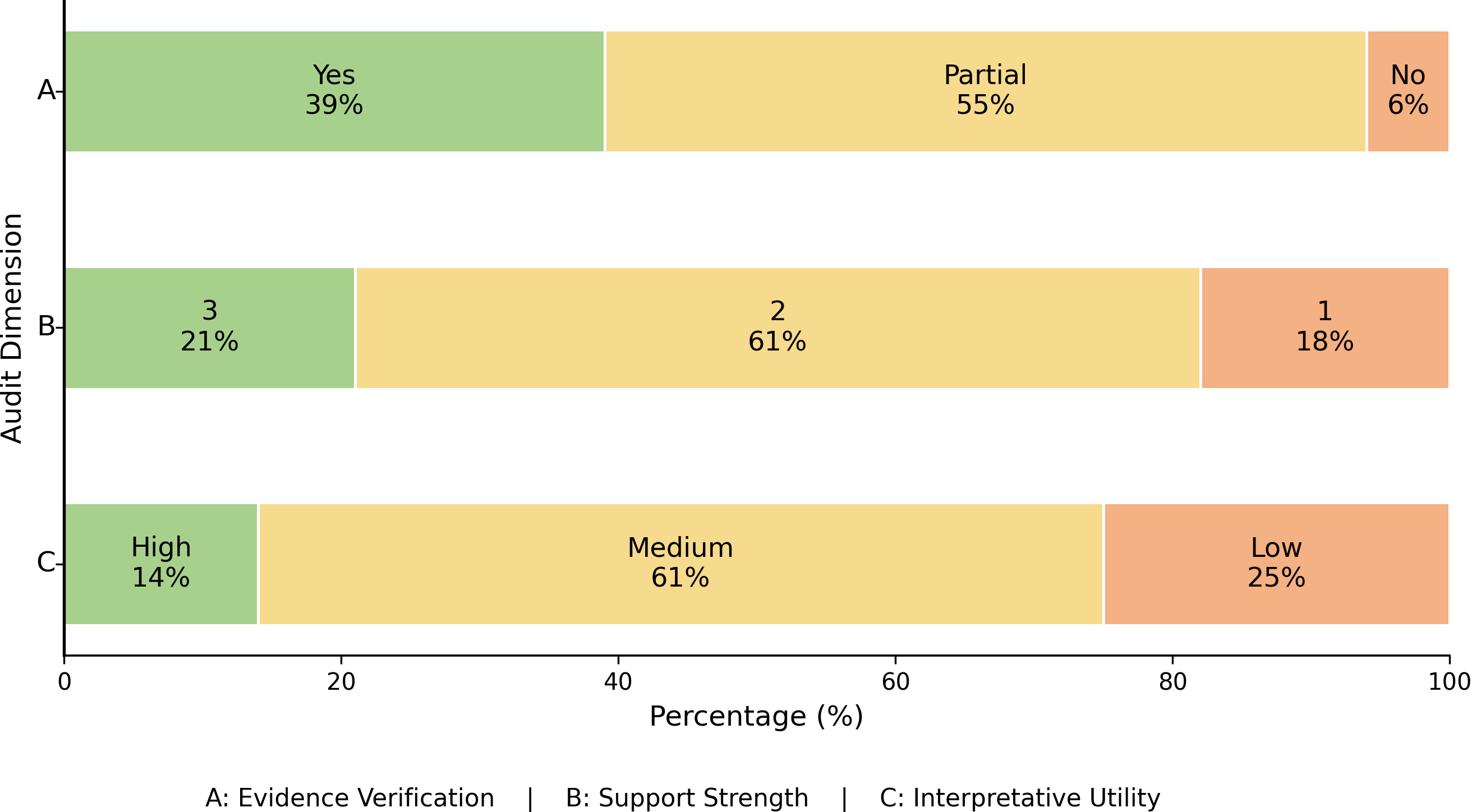}\\
  \caption{Human assessment of reviewed claims. (A) Yes: clearly present, Partial: partially present, No: absent or contradictory; (B) 1: weak, 2: acceptable, 3: strong; and (C) High: explicit and clear, Medium: requiring interpretation, Low: vague or missing link.}
  \label{fig: Audit}
  \end{center}
\end{figure}

As shown in Fig.~\ref{fig: Audit}, 39 claims are fully verifiable, 55 were partially verifiable, and 6 are not verifiable due to missing citation or source evidence. Because evaluator judgements vary in some cases, categorical labels, including evidence verification and interpretative utility, are finalised through adjudication, while support strength is summarised using the median score and checked for consistency.

In the support strength assessment, 21 claims receive strong support, while the most claims are at least partially supported and only 18 claims have weak evidential support. Interpretative utility followed a similar pattern, with 14 high-utility, 61 medium-utility, and 25 low-utility claims. 

The audit also showed that main research-question responses had stronger provenance than follow-up responses. Among the audited main-response rows, 10 of 11 were fully verifiable. In contrast, 56 of 89 follow-up rows were only partially verifiable, mainly because several follow-up rows exported the full retrieved chunk rather than a short exact evidence phrase. Thus, the evidence trail remained useful for human inspection, but follow-up responses often required extra manual checking to identify the precise supporting phrase.

\label{sec:human-provenance-audit}

\label{tab:human-audit-summary}

\subsection{Discussion}

The triangulated evaluation shows that DysLexLens can generate reasonably accurate and inspectable responses from dyslexia-related Reddit data. RAGAS benchmarking indicates strong \textit{Answer Relevancy}, particularly for the research questions, but medium \textit{Context Relevancy} which is justifiable due to the low-resource data. Query robustness analysis shows that the framework is reasonably stable under paraphrasing but more sensitive to keyword perturbation, while the human-grounded audit shows the risk of hallucination which confirms the need for claim-level expert verification. 

Nevertheless, DysLexLens can still be useful as an evidence-traceable exploratory analysis tool that supports research on dyslexic learners’ lived experiences with AI, assistive technologies, and learning support, which are often difficult to capture through formal studies alone.
The retrieved subgraphs from the KG suggests that dyslexic learners perceive existing AI tools as useful but still limited, indicating scope for further AI development in this area. 

A key value of DysLexLens is that it reduces the burden of exploratory thematic analysis in low-resource research settings while supporting reproducibility through a transparent evidence-traceable design. By separating concept-based filtering, knowledge-graph construction, retrieval-augmented generation, evidence tracing, and provenance auditing, DysLexLens provides a modular and reusable pipeline for analysing low-resource online discourse. This modularity supports generalisability, as domain experts can replace the dyslexia-AI concept dictionary with a context-specific dictionary in the concept-based filtering module, redefine the questions, and apply the same retrieval, reasoning, evaluation benchmarks, and audit guidelines to assess whether the framework produces grounded and interpretable outputs in a new setting.
 
Future versions of DysLexLens should improve synonym-aware retrieval and evidence ranking to ensure interpretations are consistently supported by precise source passages. Moreover, expanding query-variant generation through AI-based query optimisation, guided by the concept dictionary, could be a useful addition to improve response quality.

\section{Conclusion} \label{sec:conclusion}
This paper presented DysLexLens, a low-resource LLM-assisted framework for analysing dyslexic learners’ experiences with AI in online forum discussions. 
The value of DysLexLens lies in its integration of concept-dictionary filtering with evidence-traceable question answering, as demonstrated through the Reddit-based case study, which provides an explainable and reproducible pipeline. The knowledge graph supported relational interpretation across AI, technology, support, perception, and educational challenges, while the Q/A pipeline generated source-grounded responses to user-defined questions. The evaluation further shows the importance of combining quantitative metrics with qualitative expert review, as automated scores alone may not fully capture weak grounding, retrieval mismatch, or unsupported interpretation. This study has limitations. Reddit discussions do not represent all dyslexic learners, and LLM-based triple extraction and retrieval may introduce noise. Therefore, the generated graph should be treated as an interpretive aid rather than a complete representation of learner experience.

Future work will extend DysLexLens beyond this dyslexia-focused Reddit case study. By indexing new datasets and developing systematic strategies for building domain-specific concept dictionaries or alternative filtering methods, this modular framework can be adapted to other Reddit communities and online data sources, such as podcasts and social media, to answer user-defined research questions. This positions DysLexLens as a general evidence-traceable analysis tool for low-resource online discourse, particularly where researchers need to explore noisy textual data, ask structured and follow-up questions, inspect supporting evidence, and evaluate response quality.

\section*{GenAI Usage Disclosure}
In preparing this paper, GPT-5.5 is used for identifying and correcting grammatical errors and typos, as well as for assisting in the generation of illustrative figures. In accordance with academic integrity, all research content, methods, data analysis, and paper drafting are developed, conducted, and validated by the authors.

\bibliographystyle{ACM-Reference-Format}
\bibliography{dyslexia}

\end{document}